\documentclass[fleqn,10pt]{wlscirep}
\usepackage{amsmath}
\usepackage{amsfonts}
\usepackage{amssymb}
\usepackage{graphicx}
\usepackage{algorithmic}
\usepackage{algorithm}
\usepackage{color}
\usepackage{authblk}
\usepackage{algorithm}
\usepackage{algorithmic}
\usepackage{booktabs}
\usepackage{multirow}
\usepackage{array}

\newcolumntype{C}[1]{>{\centering\arraybackslash}m{#1}}
\newcolumntype{L}[1]{>{\raggedright\arraybackslash}m{#1}}
\newcolumntype{V}{!{\vrule width 1.5pt}}

\definecolor{darkgreen}{rgb}{0,0.6,0.2}

\title{Multiplex visibility graphs to investigate recurrent neural network dynamics}

\author[1,*]{Filippo Maria Bianchi}
\author[2]{Lorenzo Livi}
\author[3,4]{Cesare Alippi}
\author[1]{Robert Jenssen}
\affil[1]{Machine Learning Group, Department of Physics and Technology, University of Troms\o{}, 9019 Troms\o{}, Norway}
\affil[2]{Department of Computer Science, College of Engineering, Mathematics and Physical Sciences, University of Exeter, Exeter EX4 4QF, United Kingdom}
\affil[3]{Department of Electronics, Information, and Bioengineering, Politecnico di Milano, 20133 Milan, Italy}
\affil[4]{Faculty of Informatics, Universit\`a della Svizzera Italiana, 6900 Lugano, Switzerland}
\affil[*]{filippo.m.bianchi@uit.no}


\begin{abstract}
A recurrent neural network (RNN) is a universal approximator of dynamical systems, whose performance often depends on sensitive hyperparameters. 
Tuning them properly may be difficult and, typically, based on a trial-and-error approach.
In this work, we adopt a graph-based framework to interpret and characterize internal dynamics of a class of RNNs called echo state networks (ESNs). 
We design principled unsupervised methods to derive hyperparameters configurations yielding maximal ESN performance, expressed in terms of prediction error and memory capacity.
In particular, we propose to model time series generated by each neuron activations with a horizontal visibility graph, whose topological properties have been shown to be related to the underlying system dynamics.
Successively, horizontal visibility graphs associated with all neurons become layers of a larger structure called a multiplex.
We show that topological properties of such a multiplex reflect important features of ESN dynamics that can be used to guide the tuning of its hyperparamers.
Results obtained on several benchmarks and a real-world dataset of telephone call data records show the effectiveness of the proposed methods.
\end{abstract}

\begin{document}

\flushbottom
\maketitle
\thispagestyle{empty}

\section*{Introduction}

A current research trend aims at investigating complex time-variant systems through graph theory, by considering suitable features associated with vertices and edges \cite{barzel2013universality}.
Of particular interest are those systems that also perform a computation when driven by an external input signal. 
An example is that of artificial RNNs \cite{Hammer20041061,maass2007computational,reinhart2012regularization}, which are computational dynamical systems whose link with physics and neurosciences dates back to the '80 with some pioneering works from Jordan \cite{Jordan1997471} and Amit et al. \cite{PhysRevA.32.1007}. 
Nowadays, RNNs are gaining renewed interest in neuroscience due to their biological plausibility \cite{enel2016reservoir,barak2013fixed,rajan2010stimulus,fusi2016neurons} and in computer science and engineering for their modeling ability \cite{Schmidhuber201585,barra2012equivalence}.
RNNs are capable to generate complex dynamics and perform inference based on current inputs and internal state, the latter maintaining a vanishing memory of past inputs \cite{charles2016distributed,tivno2013short}.

\begin{figure}[!ht]
\centering
  \includegraphics[width=0.6\textwidth, keepaspectratio]{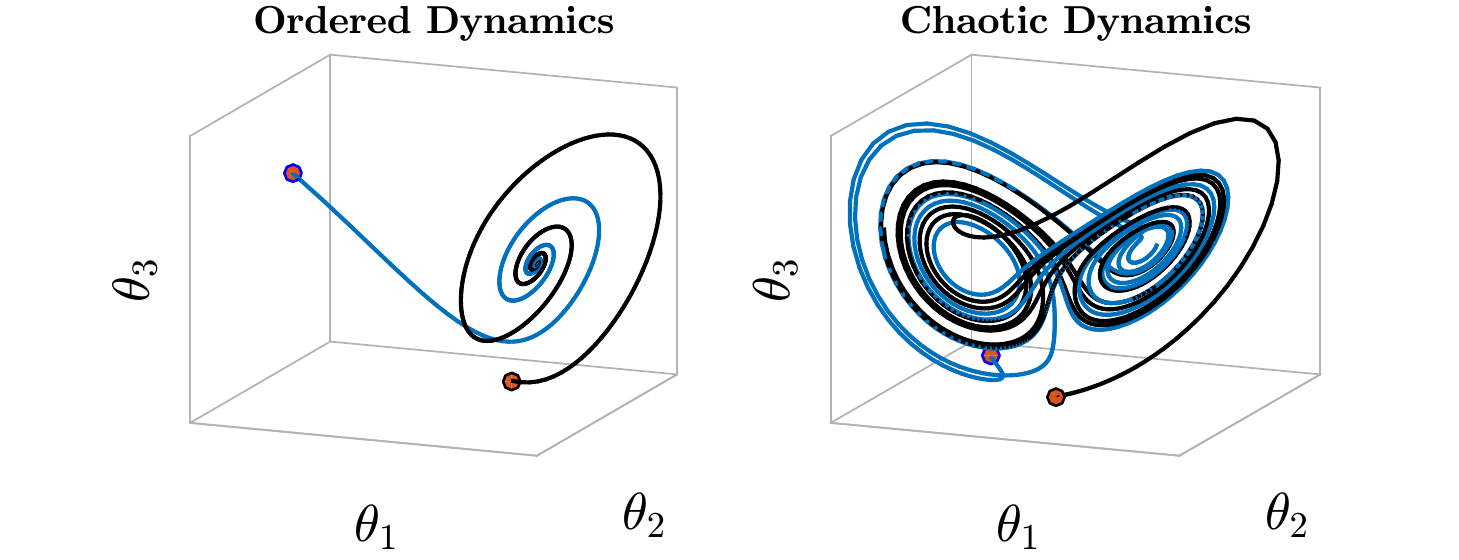}
  \caption{
Trajectories in the state space of a system described by the evolution of variables $\theta_1$, $\theta_2$, and $\theta_3$ over time. The behavior of the system is controlled by a parameter (not specified), which we assume to be able to produce ordered (left plot) and chaotic (right plot) dynamics, depending on its value. In the ordered regime, the trajectories converge to the same fixed point when starting from 2 different initial conditions (red dots). In this configuration, the system is characterized by a fading memory of previous states. In the chaotic regime, on the other hand, trajectories remain separated if the system starts from different initial conditions.}
\label{fig:attractors}
\end{figure}

Let us consider trajectories describing the evolution of a dynamical system in state space, e.g., the space containing all possible system states.
As an example, in Fig. \ref{fig:attractors} we show the trajectories of a dynamical system operating in ordered (left) and chaotic (right) regimes, whose state is defined by the values of variables $\theta_1$, $\theta_2$, and $\theta_3$ at time $t$.
%
%
Discriminating between order and chaos is of fundamental importance for investigating the properties of a dynamical system. 
It emerges that, such properties are manifested in the memory of the dynamical system and in the divergence rate of its state trajectories \cite{legenstein2007makes}. 
Fading memory is a desirable property of a dynamical system, and is characterized by ordered (contractive) dynamics. 
This is also referred to as the echo state property in the reservoir computing community \cite{yildiz2012re} and ensures that the current state/output of the system depends only on a finite number of past states/inputs \cite{dambre2012information}. 
At the same time, a high divergence rate between state trajectories (a property of chaotic dynamics) is also a desirable feature of RNNs. 
Results show that RNNs operating in a chaotic regime are able to produce meaningful patterns of activity \cite{10.1371/journal.pcbi.1005258} and a balance must be struck in order to meet both properties. 
As a consequence, a (computational) dynamical system has to operate on the transition between order and chaos, in a region of the controlling parameter space called ``edge of criticality''. On the edge of criticality, RNN internal dynamics becomes richer, meaning that neuron activations become heterogeneous \cite{legenstein2007makes}. Such a diversity both improves memory capacity of RNNs as well as their capability to reproduce complex target dynamics \cite{esnfish2016,mayer2015input}. The notion of edge of criticality permeates several complex systems \cite{moretti2013griffiths,mora2011biological,scheffer2012anticipating}, including random Boolean networks \cite{6792593} and populations of spiking neurons \cite{tkavcik2015thermodynamics}. It is furthermore well-known that several complex systems spontaneously adapt to operate towards the edge of criticality, according to a mechanism known as self-organizing criticality in the statistical physics literature \cite{markovic2014power}.

In this work, we study ESNs \cite{lukovsevivcius2009reservoir,grigoryeva2015optimal}, a class of RNNs characterized by a large, untrained recurrent layer producing outputs by means of a linear (trained) readout layer.
As such, ESNs are trained by optimizing readout weights only through a fast linear regression, rather than learning the weights of the recurrent layer by means of backpropagation through time, as in more general RNNs \cite{Schmidhuber201585}.
Please, note that more sophisticated readout layers, requiring longer training procedures, could be used as well \cite{bianchi2015prediction}.
ESNs trade the precision of gradient descent with the ``brute force'' redundancy of the random recurrent layer. This, inevitably, makes ESNs more sensitive to selection of the hyperparameters that control their internal dynamics.
Therefore, the difficulty involved in finding the optimal hyperparameters is more crucial for ESNs than in other RNN models.
The non-linear recurrent nature of ESNs hampers interpretability of internal state dynamics \cite{bianchi2016investigating}. As such, ESNs (as any other RNN) are usually treated as \textit{black box} models with hyperparameters tuned in a supervised manner through cross-validation. However, cross-validation requires evaluation of the computational performance on a validation set for each hyperparameter configuration. This might be a serious limitation in real-life applications, whenever the data are scarce and/or supervised information unavailable. Moreover, the model has to be re-trained for each hyperparameter configuration; this is an issue whenever the learning procedure is computationally demanding \cite{bianchi2015prediction}.

Several unsupervised methods for ESNs tuning have been proposed in the literature, e.g., see Refs. \cite{boedecker2012information,ozturk2007analysis}.
Although unsupervised tuning seldom achieves the same level of accuracy of the supervised counterpart, it offers some insights to the interpretation of the system behavior \cite{verstraeten2009quantification}.
Typically, unsupervised learning is achieved by observing the dynamics of the states.
These methods are usually based on a statistical framework, which requires reservoir outputs to be independent and identically distributed (i.i.d.).
However, internal states are not identically distributed when ESNs are driven by non-stationary inputs \cite{NIPS1998_1529}. 
Independence, instead, is always violated both by dependent inputs and by temporal dependencies introduced by recurrent connections.
Therefore, even if the input signal is i.i.d., neurons activations may be identically distributed but never independent.

With this work, we propose two unsupervised graph-based methods for tuning ESN hyperparameters, with the aim to maximize (i) prediction accuracy and (ii) memory capacity.
In our framework, temporal dependencies in time series of neurons activations (states) are converted into connections of a graph representation.
This step allows to relax the i.i.d. assumption of statistical methods. 
Then, graphs associated with all neurons activations become layers of a structure called a multiplex.
A multiplex graph \cite{lee2015towards,boccaletti2014structure,menichetti2014weighted,kivela2014multilayer,de2013mathematical} is a special type of multilayer network, whose vertices are replicated through each layer and connected across layers only with their replicas.
The topology (i.e., the edges) of the graph in each layer can be different.
In particular, here we use a multiplex visibility graph \cite{lacasa2015network}, whose layers consist of a particular class of graphs called horizontal visibility graphs (HVGs) \cite{PhysRevE.80.046103}.
HVGs are planar graphs built from real-valued time series, whose elements have one-to-one correspondence with the vertices.
Important properties, linking the structure of HVGs with features (e.g., onset of chaotic behavior) of the dynamic system underlying the analyzed time series, have been recently studied \cite{luque2011feigenbaum,luque2013quasiperiodic,lacasa2010description,ravetti2014distinguishing,luque2012analytical}. This last aspect provides an important justification for using HVGs to model neuron activations for the purpose of hyperparameter tuning.

The novelty introduced in our paper refers to the characterization of ESN dynamics through structural characteristics of the associated multiplex network.
To the best of our knowledge, this is the first time ESNs are studied within the framework of multiplex networks.
However, it is worth citing a loosely related paper by Zhang et al. \cite{zhang2016architectural}, where the authors set the basis for a graph-theoretical analysis of RNNs.
Here, we represent the instantaneous state of an ESN through a set of vertex properties of the multiplex (e.g., degree, clustering coefficient). 
Edges in HVGs might cover a relevant time interval (contained within the longest period of the signal), while they are still local in terms of topology. 
Accordingly, the ESN state can be characterized also by information that is non-local in time.
To find hyperparameters yielding highest prediction accuracy, we search hyperparameter configurations producing neuron activations as diverse as possible.
This occurs when neurons dynamics is maximally heterogeneous (critical dynamics), a characteristic that, as we will show in the paper, is well-captured by the average entropy of vertex properties of the multiplex.
Successively, to quantify the amount of memory in an ESN, we check the existence of neuron activations that are ``similar'' to different delayed versions of the input.
In fact, memory in ESNs depends on the ability to reproduce past input sequences from information kept within some neuron activations.
We describe dynamics of delayed inputs and of neuron activations through unsupervised graph-based measures. 
Then, by evaluating the agreement between such measurements, we quantify the memory capacity.
We provide experimental evidence that our methods achieve performance comparable with supervised techniques for identifying hyperparamer configurations with high prediction accuracy and large memory capacity.

\section*{Methods}
\label{sec:methods}

\subsection*{Echo state networks}
\label{sec:esn}

A standard ESN architecture consists of a large, recurrent layer of non-linear neurons sparsely interconnected by edges with randomly generated weights, called a reservoir, and a linear, feedforward readout layer that is usually trained with regularized least-squares optimization \cite{lukovsevivcius2009reservoir}.
De facto, the recurrent layer acts as a non-linear kernel \cite{hermans2012recurrent} that maps inputs to a high-dimensional space.
The time-invariant difference equations describing the ESN state-update and output are, respectively, defined as
\begin{align}
\label{eq:state_update}
\mathbf{h}[t] =& f(\mathbf{W}_{r}^{r}\mathbf{h}[t-1] + \mathbf{W}_{i}^{r}\mathbf{x}[t] + \mathbf{W}_{o}^{r}\mathbf{y}[t-1]),\\
\label{eq:esn_output}
\mathbf{y}[t] =& g(\mathbf{W}_{i}^{o}\mathbf{x}[t] + \mathbf{W}_{r}^{o}\mathbf{h}[t]),
\end{align}

The reservoir contains $N_r$ neurons whose transfer/activation function $f(\cdot)$ is usually a hyperbolic tangent and $g(\cdot)$ is usually the identity function.
At time instant $t$, the ESN is driven by the input signal $\mathbf{x}[t]\in \mathbb{R}^{N_i}$ and produces the output $\mathbf{y}[t] \in \mathbb{R}^{N_o}$, being $N_i$ and $N_o$ input and output dimensions, respectively.
The vector $\mathbf{h}[t] \in \mathbb{R}^{N_r}$ describes the ESN (instantaneous) state.
The weight matrices $\mathbf{W}_r^r \in \mathbb{R}^{N_r \times N_r}$ (reservoir connections), $\mathbf{W}_i^r \in \mathbb{R}^{N_i \times N_r}$ (input-to-reservoir connections), and $\mathbf{W}_o^r \in \mathbb{R}^{N_o \times N_r}$ (output-to-reservoir connections) contain real values randomly sampled from a uniform or Gaussian distribution.
Weights in $\mathbf{W}_{i}^{o}$ and $\mathbf{W}_{r}^{o}$, instead, are trained (typically by means of regularized least-squares algorithm) according to the task at hand.

Once given, the designer can control $\mathbf{W}_{r}^{r}$, $\mathbf{W}_{o}^{r}$, and $\mathbf{W}_{i}^{r}$ only through scaling coefficients.
For instance $\mathbf{W}_{o}^{r}$ is scaled by a multiplicative constant $\omega_{o}$, which has the effect to tune the impact of the ESN output on the next state.
In this study, the output feedback is removed by setting $\omega_{o} = 0$. Hence, the resulting state-update difference equation (\ref{eq:state_update}) becomes
\begin{equation}
    \label{eq:esn_state_nooutputfeedback}
    \mathbf{h}[t] = f(\mathbf{W}_{r}^{r}\mathbf{h}[t-1] + \mathbf{W}_{i}^{r}\mathbf{u}[t]),
\end{equation}
while the output equation (\ref{eq:esn_output}) remains unchanged.

The properties of the ESN described in (\ref{eq:esn_state_nooutputfeedback}) heavily depend on two hyperparameters. First, input weights in $\mathbf{W}_i^r$ are scaled by a multiplicative constant $\omega_i$, which controls the amount of non-linearity introduced by neurons. Large values for $\omega_i$ tend to saturate the non-linear activation functions. The second hyperparameter is the spectral radius $\rho$ (eigenvalue with largest absolute value) of $\mathbf{W}_r^r$, which is related to the echo-state property, discussed earlier.  For a detailed discussion on the relationships between stability, performance and $\rho$, we suggest to the interested reader Ref. \cite{bianchi2016investigating} and references therein. Here, it suffices to say that a widely adopted rule-of-thumb \cite{Lukosevicius2012} suggests to set $\rho$ to a value slightly smaller than 1 (e.g., $\rho = 0.99$). However, to reach higher performance in some practical tasks, it could be necessary to pick a small value for $\rho$ or to breach the aforementioned ``safety'' bound and push its value beyond unity. Note that in this latter case asymptotic stability of the ESN might still hold, even if some assumptions are locally violated.

In this work, we focus only on the tuning of the hyperparamers $\rho$ and $\omega_i$, without affecting generality of the proposed methods that can be adopted to tune also other hyperparamers.
The optimal values of $\rho$ and $\omega_i$ for the task at hand are typically identified with a cross-validation procedure, with the potential associated shortcomings mentioned above. For that reason, different unsupervised approaches have been proposed for their tuning \cite{boedecker2012information,bianchi2016investigating,verstraeten2009quantification}.
For example, the effect of $\rho$ and $\omega_i$ on the ESN computational capability can be investigated through the maximal local Lyapunov exponent (MLLE), which measures the divergence rate in state space of trajectories with similar initial conditions.
In autonomous (not input-driven) systems, chaos occurs when the maximal Lyapuanov exponent becomes positive, while in input-driven systems, like ESN, one typically relies on local first-order approximations of this quantity (see \cite{bianchi2016investigating} for details).
Accordingly, the onset of criticality in ESNs can be detected by checking when MLLE crosses 0.
Another quantity, which was shown to be more accurate in detecting criticality in dynamic systems and well-correlated with ESN performance, is the minimal singular value (in average, over time) of the reservoir Jacobian, denoted as $\lambda$ \cite{verstraeten2009quantification}. 
$\lambda$ is unimodal and in correspondence of its maximum the dynamical system is far from singularity, has many degrees of freedom, has a good excitability, and it separates well the input signals in state space \cite{verstraeten2009quantification}.
By assuming a null input in Eq. \ref{eq:esn_state_nooutputfeedback}, the Jacobian matrix of the reservoir at time $t$ is given by
$\mathbf{J}_t = \mathrm{diag}(1 - (\mathbf{h}_1[t])^2, 1 - (\mathbf{h}_2[t])^2, \dots , 1 - (\mathbf{h}_{N_r}[t])^2) \mathbf{W}_{r}^{r}$, where the $\mathrm{diag}(\cdot)$ operator returns a diagonal matrix.
In this paper, $\lambda$ will serve as a baseline for comparison to our proposed graph-based unsupervised methods for improved ESN hyperparameter tuning, as discussed in the next section.

\subsection*{Horizontal visibility graph and multiplex network}

The HVG \cite{PhysRevE.80.046103} associated with a finite univariate time series $\mathbf{x} = \left\{\mathbf{x}[t]\right\}_{t=1}^{t_\text{max}}$, is constructed by assigning a vertex $v_t$ to each datum $\mathbf{x}[t]$.
The adjacency matrix $\mathbf{A}$ characterizes the graph: two vertices $v_i$ and $v_j$, $i\neq j$ are connected by an edge ($\mathbf{A}[i,j] = 1$) iff the corresponding data fulfill the criterion $\mathbf{x}[t_i], \mathbf{x}[t_j] > \mathbf{x}[t_p], \forall \; t_i < t_p < t_j$.
In a multivariate scenario, the data stream is composed of $N_r$ different time series $\{ \mathbf{x}_l\}_{l=1}^{N_r}$, of equal length $t_\text{max}$.
A multivariate time series can be mapped into a multiplex visibility graph $\mathcal{M}$ with $N_r$ layers \cite{lacasa2015network}.
Specifically, the $l$th multiplex layer is defined by the HVG $G_l$ constructed from $\mathbf{x}_l$.
In the multiplex, a vertex is replicated on all layers and such replicas are linked by inter-layer connections, while intra-layer connections might change in each layer. From now on, we denote $\mathbf{v}_l[t]$ to be the vertex of $G_l$ in layer $l$ associated with time interval $t$.

In this paper we introduce a weighted HVG (wHVG), with edge values defined as $\mathbf{A}[i,j] = 1/\sqrt{(j-i)^2 + (\mathbf{x}[i]-\mathbf{x}[j])^2}\in[0, 1], \forall 1\leq i, j \leq t_\text{max}$. Since self-loops are forbidden in HVGs, edge weights are always well-defined (i.e., finite). 
The use of weights permits to capture additional information, as it accounts for distance in time $(j-i)$ and amplitude differences $(\mathbf{x}[i]-\mathbf{x}[j])$ of two data points connected by the visibility rule. 
This weighting scheme is motivated by our need to characterize and exploit the instantaneous state by means of a suitable measure of heterogeneity (discussed in the next section). 
To distinguish the original adjacency matrix from the one of the wHVG, we refer the former as \textit{binary} adjacency matrix and the latter as weighted adjacency matrix.
Algorithm \ref{alg:hvg} delivers the pseudo-code for constructing a HVG (and a wHVG) from time series $\mathbf{x}$.
The worst case complexity of this algorithm is $\mathcal{O}(t_\text{max}^2)$, which occurs when values in $\mathbf{x}$ are monotonically decreasing. Instead, the best case complexity is ${\scriptstyle\mathcal{O}}(t_\text{max})$ and arises in correspondence of monotonically increasing values in $\mathbf{x}$.
%
\begin{algorithm}[!ht]\scriptsize
\caption{Construction of a (weighted) HVG.}
\label{alg:hvg}
\begin{algorithmic}[1]
\REQUIRE Time series $\left\{\mathbf{x}[t]\right\}_{t=1}^{t_\text{max}}$
\ENSURE Adjacency matrix $\mathbf{A}$ of (weighted) HVG
  \FOR{$i=1, ..., t_\text{max}-1$}
    \STATE Set $j=i+1$, $\mathrm{max=-\infty}$, $\mathrm{stop=false}$, $\mathrm{count}=0$
    \WHILE{stop is false AND $j\leq t_\text{max}$}
      \STATE $\mathrm{count=count}+1$
      \IF{$\mathbf{x}[j]>\mathrm{max}$}
	\STATE If unweighted, $\mathbf{A}[i,j]=1$, otherwise, $\mathbf{A}[i,j]=1/\sqrt{(j-i)^2 + (\mathbf{x}[i]-\mathbf{x}[j])^2}$
	\STATE $\mathrm{max}=\mathbf{x}[j]$
	\IF{$\mathrm{max}>\mathbf{x}[i]$}
	  \STATE $\mathrm{stop=true}$
	\ENDIF
      \ENDIF
      \STATE $j=j+1$
    \ENDWHILE
  \ENDFOR
\end{algorithmic}
\end{algorithm}

\subsection*{Vertex properties}
\label{sec:vertex_prop}

Let us consider an ESN with a reservoir of size $N_r$, driven by an input of length $t_\text{max}$.
According to (\ref{eq:esn_state_nooutputfeedback}), the ESN generates $N_r$ time series $\left\{ \mathbf{h}_1[t] \right\}_{t=1}^{t_\text{max}}, \dots, \left\{ \mathbf{h}_{N_r}[t] \right\}_{t=1}^{t_\text{max}}$, of neuron activations (state).
The multivariate time series $\left\{ \mathbf{h}_l \right\}_{l=1}^{N_r}$ is represented here by a multiplex $\mathcal{M}$.
Indexes of the vertices on each layer $l$ of $\mathcal{M}$ are associated one-to-one with the time index of the original time series.
Hence, the ESN state $\{ \mathbf{h}_l[t] \}_{l=1}^{N_r}$ at time $t$ is represented by the vertices $\left\{ \mathbf{v}_l[t] \right\}_{l=1}^{N_r} $ and by vertex properties $\Phi^*[t] = \{ \phi^* \left( \mathbf{v}_l[t] \right) \}_{l=1}^{N_r}$.
In Tab. \ref{tab:vertex_char} we introduce four indexes: vertex degree, clustering coefficient \cite{saramaki2007generalizations}, betweenness and closeness centrality \cite{boccaletti+latora+moreno+chavez+hwang2006}.

\bgroup
\def\arraystretch{1.4} 
\setlength\tabcolsep{.2em} 
\begin{center}
\begin{table}[!ht]\scriptsize
  \caption{Vertex characteristics used to generate different instances of the vertex property vector $\boldsymbol{\phi}_t$.}
  \begin{tabular}{VC{1.4cm}|L{2.8cm}L{10cm}V}
    \noalign{\hrule height 1.5pt}
    Vertex degree & \hspace{0.3cm} $\boldsymbol{\phi}^{\mathrm{DG}}(v) = \sum_i \mathbf{A}[i,v]$  & Number of edges incident to vertex $v$. \\
    \noalign{\hrule height .5pt}
    Clustering coefficient & \hspace{0.3cm} $\boldsymbol{\phi}^{\mathrm{CL}}(v) = \frac{ \sum \limits_{i \in \mathcal{C}_v} \sum \limits_{j \in \mathcal{C}_v} \mathbf{A}[i,j] }{|\mathcal{C}_v|\cdot\left( |\mathcal{C}_v| - 1\right)}$ & Clustering coefficient of vertex $v$, applicable to both weighted and binary graphs. In Ref.\cite{lopez2004applying}, $\mathcal{C}_v = \left\{ i | \mathbf{A}[i,v] \neq 0 \vee \mathbf{A}[v,i] \neq 0 \right\}$. Here, we consider $\mathcal{C}_v = \left\{ v \cup \mathcal{C}_v \right\}$ to include also $v$ (and its edges). \\
    \noalign{\hrule height .5pt}
    Betweenness Centrality & \hspace{0.3cm} $\boldsymbol{\phi}^{\mathrm{BC}}(v) = \displaystyle\sum_{i\neq v \neq j} \frac{\sigma_{ij}(v) }{\sigma_{ij}}$ & Measures the centrality of vertex $v$. $\sigma_{ij}$ is the total number of shortest paths from $i$ to $j$ and $\sigma_{ij}(v)$ denotes the number of shortest paths passing through $v$. \\
    \noalign{\hrule height .5pt}
    Closeness centrality & \hspace{0.3cm} $\boldsymbol{\phi}^{\mathrm{CC}}(v) = \displaystyle\sum_{v \neq i} 2^{-|\sigma(i, v)|}$ & Total distance of a vertex $v$ from all other vertices in the graph. $|\sigma(i, v)|$ denotes the length of the shortest path between $i$ and $v$ \\
    \noalign{\hrule height 1.5pt}
  \end{tabular}
  \label{tab:vertex_char}
\end{table}
\end{center}
\egroup

\subsection*{Heterogeneity of neurons dynamics}

The capability of an ESN to reproduce the dynamics of a target system, hence to predict its trajectory in state space, is maximized on the edge of criticality, where the internal dynamical patterns of an ESN become sufficiently rich.
In the literature, such a ``richness'' is usually expressed in terms of diversity of connection weights \cite{bertschinger2004real}, entropy or rank of the matrix of neuron activations \cite{ozturk2007analysis,legenstein2007makes}.

\begin{figure}[!ht]
\centering
  \includegraphics[width=0.9\textwidth, keepaspectratio]{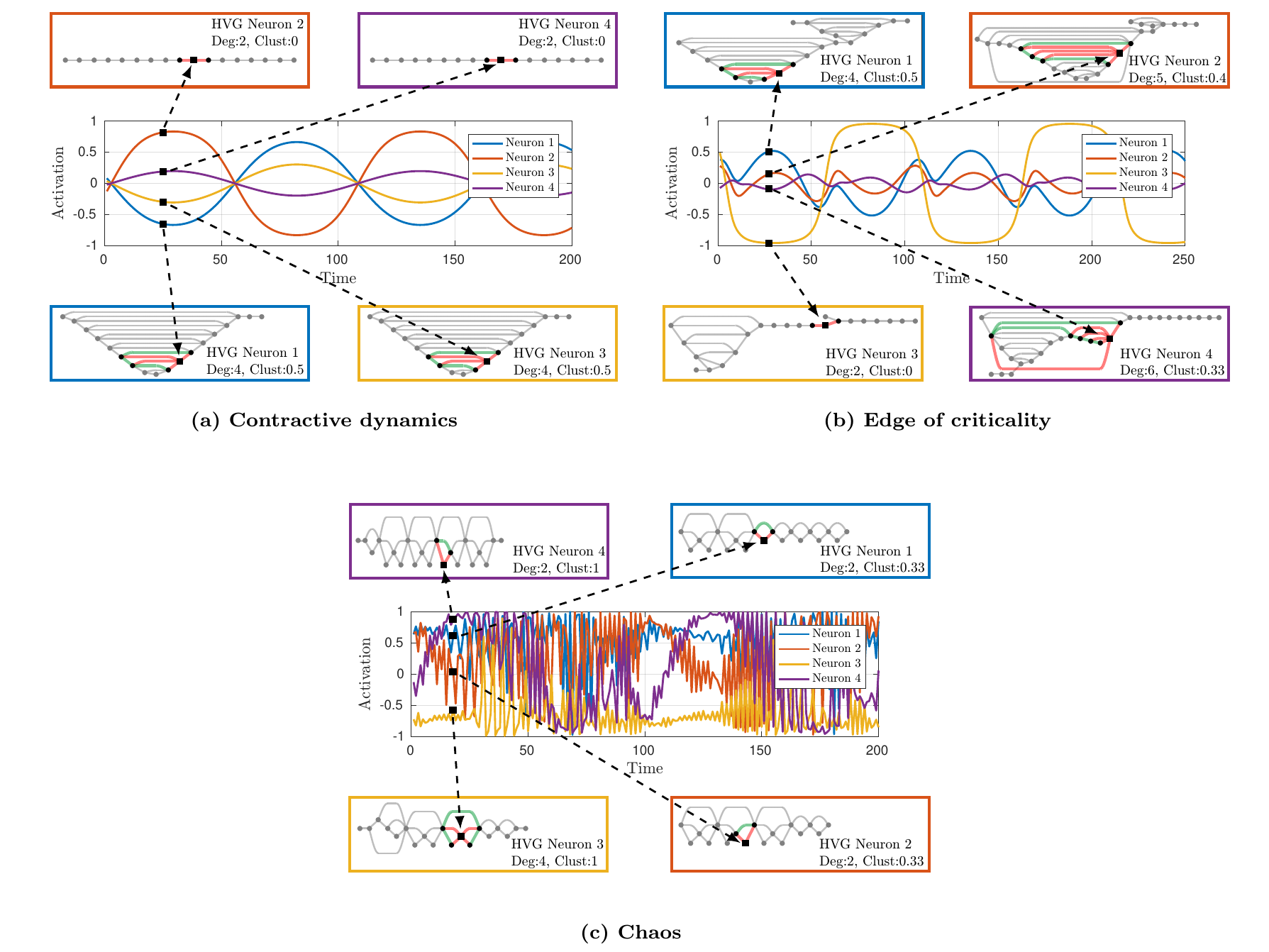}
  \caption{Heterogeneity of the graph-based representation of the ESN instantaneous state. From each time series, we visualize only a portion of its HVG and consider a specific time step (black square). In the HVGs, we highlight in red the edges connecting each vertex that corresponds to the selected time step. For the clustering coefficient, instead, one has to consider also edges highlighted in green. The non-local nature of ESN graph-based state is evident from such representations, since vertices can be connected to other vertices that are associated to time instants far away along the sequence.}
\label{fig:heterogeneity}
\end{figure}

In the same spirit, in order to find hyperparameters yielding maximal prediction accuracy, here we look for those hyperparameter configurations giving rise to neuron activations that are as heterogeneous as possible.
Fig. \ref{fig:heterogeneity} provides a visual example of the concept herein discussed. 
As an illustration, we consider an ESN driven by a sinusoid. We depict the neuron activations and the corresponding HVGs.
We select a time step (marked by a black square in the picture) and we show the correspondence between the element in each time series of the neuron activations and the vertices in the associated HVGs.
%
%
When the ESN operates with a contractive dynamic (\ref{fig:heterogeneity}a), neuron activations weakly depend on previous ESN states and they are all very similar to the input signal. 
This results in a lack of diversity among activations. 
Accordingly, the corresponding HVGs in the multiplex contain vertices with similar properties, e.g., similar degree and clustering coefficient.
When the ESN approaches the edge of criticality, neuron activities highly depend on previous internal states, which encode information of past inputs and internal structure of neuron connections.
As we can see from (\ref{fig:heterogeneity}b), on the edge of criticality, the activations have different frequencies and their phases are shifted.
Such a heterogeneity in the ESN instantaneous state is captured by vertex properties of the multiplex.
Finally, if pushed beyond the edge of criticality, the ESN transits into a chaotic regime as shown in (\ref{fig:heterogeneity}c).
In this case, neuron activations become noise-like and disordered oscillations generate HVGs with vertex properties very different from previous configurations. 
However, the diversity of the patterns in different time series disappears, hence heterogeneity is again lost.
This lack of variety is also highlighted by recurrence of similar motifs in the corresponding HVGs.

To determine the heterogeneity of neuron activations, we consider the entropy of the related vertex property distribution.
In particular, heterogeneity is computed as follows:
\begin{enumerate}
 \item At each time $t$, evaluate the vertex properties $\Phi^*[t] = \{ \phi^* \left( \mathbf{v}_l[t] \right) \}_{l=1}^{N_r}$ using one of the properties listed in Tab. \ref{tab:vertex_char}.
 \item Estimate distribution $p\left(\Phi^*[t] \right)$ using a histogram with $b$ bins.
 \item Compute instantaneous entropy: $H_t = H\left(p\left(\Phi^*[t] \right)\right)$, where $H(\cdot)$ is the Shannon entropy.
 \item Define heterogeneity as average entropy over time instants: $\bar{H} = \frac{1}{t_\text{max}} \sum \limits_{t=1}^{t_\text{max}} H_t$.
\end{enumerate}

\begin{figure}[!ht]
\centering
  \includegraphics[width=0.9\textwidth,keepaspectratio]{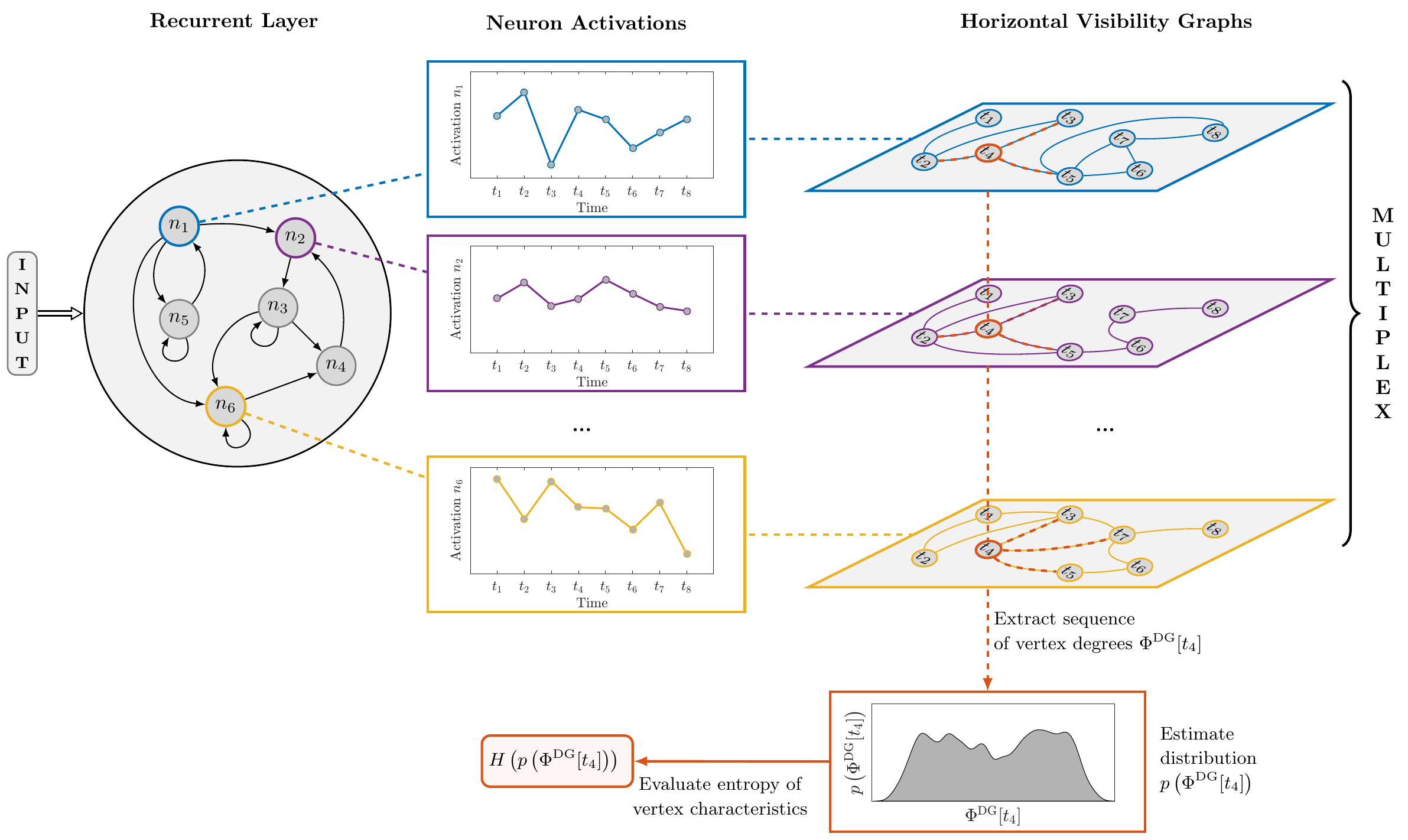}
  \caption{High-level illustration of the proposed methodology.
  The activity of an ESN that is driven by some input signal is analyzed through a multivariate time series of neuron activations.
  We map such time series with a multiplex formed by HVGs on the layers. 
  In this way, we get rid of problems related to stationarity and dependence of neuron activations. 
  Note that with such a representation, the state of an ESN is defined through graph-theoretical tools, which allows to include non-local (in time) information in the ESN instantaneous state. For instance, at time instant $t_4$, instead of the original activations, we consider a vertex property (e.g., vertex degrees $\Phi^\mathrm{DG}[t_4]$) of the HVG to represent the ESN state. The instantaneous state is eventually characterized by a measure of heterogeneity of such values; here, entropy is taken into account.}
\label{fig:schema}
\end{figure}
%
The procedure described above, is visually represented in Fig. \ref{fig:schema}.
By referring to the figure, we observe that, at given time $t$, the ESN state is represented by those vertices in the multiplex tagged with the same label across the layers.
For example, the ones in red describe the instantaneous state of the ESN at time step $t_4$ and a vector of vertex properties $\Phi^*[t_4]$ is computed for this set of vertices.
Then, we estimate the distribution $p(\Phi^*[t_4])$ and compute the entropy $H^*_4 = H\left(p(\Phi^*[t_4])\right)$.
In order to compute $\bar{H}^*$, the procedure is repeated for each time step. Finally, $\bar{H}^*$ is used to characterize the current hyperparameter configuration.
The average entropy depends on the specific vertex property chosen for the analysis (we taken into account four properties, e.g., vertex degree, clustering coefficient, betweenness and closeness centrality -- see Tab. \ref{tab:vertex_char}).
These vertex properties lead to four different entropy values $\bar{H}^\text{DG}$, $\bar{H}^\text{CL}$, $\bar{H}^\text{BC}$, and $\bar{H}^\text{CC}$.

To summarize, given an input signal, we select hyperparameter configurations that maximize such entropy values. This criterion is inspired by the aforementioned observation linking performance of a computational dynamical system (i.e., prediction accuracy and memory) with heterogeneity of its dynamics (critical dynamics).
Accordingly, in this case this information is exploited in order to derive, in an unsupervised way, the configuration yielding highest prediction accuracy.

\subsection*{Other multiplex complexity measures}
\label{sec:other_multiplex_measures}

Recently \cite{lacasa2015network}, two measures have been proposed in order to characterize the dynamics of a system observed through a multivariate time series and represented as a multiplex composed of HVGs.
Here, we consider these measures in order to evaluate whether they are useful for identifying the hyperparameter configurations yielding maximum accuracy performances or not.

The Average Edge Overlap (AEO) computes the expected number of layers of the multiplex on which an edge is present.
For binary HVGs, it is defined as
\begin{equation}
 \mathrm{AEO} = \frac{\sum_i \sum_{j>i} \sum_l \mathbf{A}_l[i,j] }{N_r \sum_i \sum_{j>i} \left( 1 - \delta_{0, \sum_l \mathbf{A}_l[i,j]} \right) },
\end{equation}
where $\mathbf{A}_l[i,j]$ equals 1 if vertices $i$ and $j$ are connected in layer $l$.
The second measure is the Interlayer Mutual Information (IMI), which quantifies the correlations between the degree distributions of two different layers $l_i$ and $l_j$. It is defined, for binary HVGs, as
\begin{equation}
 \mathrm{IMI}(l_i,l_j) = \sum \limits_{k_{l_i}} \sum \limits_{k_{l_j}} p\left(k_{l_i}, k_{l_j}\right) \log\left(\frac{p\left(k_{l_i}, k_{l_j}\right)}{p\left(k_{l_i}\right) p\left(k_{l_j}\right)}\right),
\end{equation}
where $p(k_{l_i}, k_{l_j})$ is the joint probability to find a vertex with degree $k_{l_i}$ in layer $l_i$ and degree $k_{l_j}$ in layer $l_j$, respectively.
In the experiments, we use the average IMI between all pairs of multiplex layers.

\subsection*{Memory measures}
\label{sec:memory_measures}

The memory of an ESN is quantified by the capability to retain information about past inputs in its transient dynamics \cite{pascanu2011neurodynamical}.
A supervised measure called memory capacity (MC) is usually adopted to quantify ESN memory \cite{lukovsevivcius2009reservoir}.
When computing MC, the reservoir topology and weights are kept fixed and several readout layers are trained in order to reproduce delayed versions of the input at different time lags $\tau_{L_1}, \dots, \tau_{L_\mathrm{max}}$. 
The MC is computed as the squared correlation coefficient between desired (delayed input) and computed outputs,
\begin{equation}
\label{eq:MC}
\mathrm{MC} = \displaystyle\sum \limits_{\tau_L} \text{cov}^2\left( \mathbf{x}[t-\tau_L], \mathbf{y}[t] \right) / \big( \mathrm{var}\left(\mathbf{x}[t-\tau_L]\right) \mathrm{var}\left(\mathbf{y}[t]\right)\big).
\end{equation}
In order to properly evaluate the capability of ESNs to introduce memory through recurrent connections, $\mathbf{x}[t]$ is defined as a stationary uncorrelated noisy signal.

In the following, we propose an unsupervised graph-based method to identify hyperparameter configurations for which an ESN achieves large memory capacity.
Given the input time series $\mathbf{x} = \{ x[t] \}_{t=1}^{t_{\mathrm{max}}}$, we determine if there exists a subset of neuron activations, which is correlated with a past input sequence $\mathbf{x}_{a,b} = \{ x[t-\tau_a], \dots, x[t-\tau_b] \}$, with $\tau_a > \tau_b$ and $\tau_a, \tau_b \in [1, t_\text{max}-1]$.
Being $G_l$ the HVG representing the $l$th layer of the multiplex, $\Phi_l^{\mathrm{DG}} = \{ \phi^{\mathrm{DG}} \left( \mathbf{v}_l[t] \right) \}_{t=1}^{t_\text{max}}$ is the sequence of its vertex degrees ordered according to the time index (not to be confused with $\Phi^{\mathrm{DG}}[t] = \{ \phi^{\mathrm{DG}} \left( \mathbf{v}_l[t] \right) \}_{l=1}^{N_r}$, the degrees of vertices relative to time $t$ across the different layers).
With $G_{\mathbf{x}}$, instead, we refer to the HVG constructed over the input $\mathbf{x}[t]$, while $\Phi^{\mathrm{DG}}_{\mathbf{x}}$ is the vector of its vertex degrees.

First, we define a measure of maximum agreement between $\Phi^{\mathrm{DG}}_\mathbf{x}$ and each sequence $\Phi^{\mathrm{DG}}_l$ as $\delta^{\mathrm{DG}} = \max_{l} \kappa^* \left( \Phi^{\mathrm{DG}}_\mathbf{x}, \Phi^{\mathrm{DG}}_l \right).$
$\kappa^*(\cdot, \cdot)$ is a similarity measure between sequences: in this paper we consider the Pearson correlation $\kappa^{\mathrm{PC}}(\cdot, \cdot)$, the Spearman correlation $\kappa^{\mathrm{SC}}(\cdot, \cdot)$, and the mutual information $\kappa^{\mathrm{MI}}(\cdot, \cdot)$.

A second measure of agreement is defined on the adjacency matrix $\mathbf{Q}_l = \mathbf{A}_\mathbf{x} \wedge \mathbf{A}_l$, where $\mathbf{Q}_l[i,j] = 1 \; \text{if} \; \mathbf{A}_\mathbf{x}[i,j] = 1$ and $\mathbf{A}_l[i,j]=1$; $\mathbf{Q}_l[i,j] = 0$ otherwise.
The agreement is then computed as the largest number of edges among all possible intersection graphs, $\delta^{\mathrm{AND}} = \max_{l} |\mathbf{Q}_l|$,
where $|\cdot|$ counts the number of ones in the adjacency matrix.

Finally, one might consider the similarity between original input $\mathbf{x}$ and neuron activations $\mathbf{h}_l$, $\delta^{\mathrm{TS}} = \max_{l} \kappa^*(\mathbf{x}, \mathbf{h}_l).$
In this case, $\kappa^*(\cdot,\cdot )$ is directly evaluated on the time series rather than on the sequences of vertex degrees.
This last measure of agreement is taken into account in order to quantitatively show (in the experiments) the benefits of using HVGs for representing neuron activations.

\begin{figure}[!ht]
\centering
  \includegraphics[width=0.7\textwidth,keepaspectratio=true]{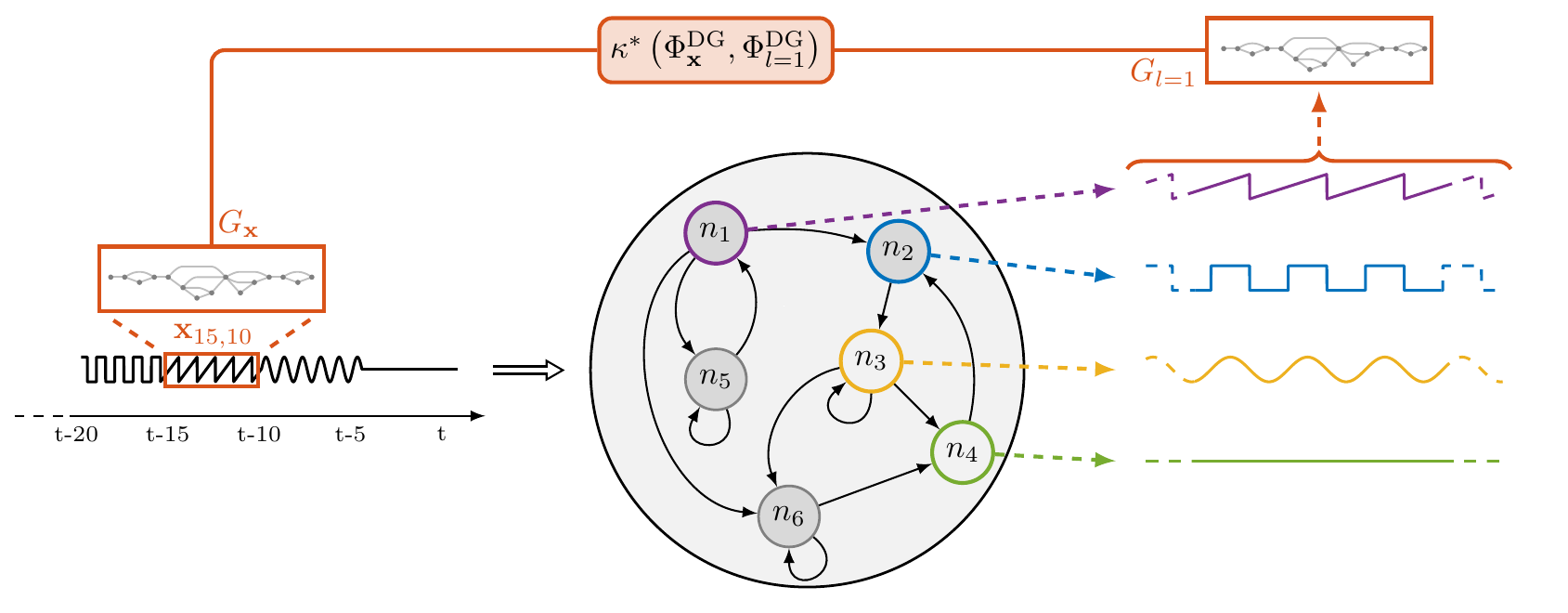}
  \caption{We extract a subsequence from the history of the input signal, for example $\mathbf{x}_{15,10}$, and compute the related HVG $G_\mathbf{x}$. 
	We repeat the procedure for each neuron activation, ending up with $G_{1}, \dots, G_{N_r}$. 
	For each HVG, we compute the sequence of vertex degrees and evaluate the similarities $\kappa^* \left( \Phi_\mathbf{x}^{\mathrm{DG}}, \Phi_{l}^{\mathrm{DG}} \right)$ for $l=1,\dots,N_r$. 
	The largest similarity value determines the value of $\delta^\text{DG}$.
	Here, the activation of neuron $n_1$ is clearly the most related one with the input sequence $\mathbf{x}_{15,10}$ and therefore it determines the value of $\delta^\text{DG}$.}
\label{fig:memory}
\end{figure}

By referring to the illustrative example in Fig. \ref{fig:memory}, we generate the HVGs $G_{\mathbf{x}}$, $G_1, \dots, G_{N_r}$, relative to delayed input $\mathbf{x}_{15,10}$ and activations of the $N_r$ reservoir neurons, respectively.
Successively, we evaluate their similarities by means of an agreement measure ($\delta^\text{DG}$ in this example). 
The similarity measure $\kappa^*$ is chosen among the three previously proposed measures: $\kappa^{\mathrm{PC}}$, $\kappa^{\mathrm{SC}}$, or $\kappa^{\mathrm{MI}}$.
%

\section*{Results}
\label{sec:results}

In the following, we perform two experiments in order to evaluate the two proposed unsupervised methods for, respectively, finding hyperparameter configurations giving rise to ESNs with high prediction accuracy and large memory capacity.
In the first experiment, we show that, on different real and synthetic tasks, (supervised) prediction accuracy is maximized for the same hyperparameter configurations that yields the largest heterogeneity for the vertex properties of the multiplex. 
In the second experiment, we show the reliability of the graph-based memory measures in identifying hyperparameters where the (supervised) MC is maximized.

\subsection*{Test for prediction accuracy}
\label{sec:test_accuracy}

In this experiment, we consider several prediction tasks and, for each of them, we set the forecast step $\tau_f>0$ to be the smallest time-lag that guarantees the measurements in a time window of size $\tau_f$ to be uncorrelated (e.g., the first zero in the autocorrelation function of the input signal).
Prediction error is evaluated by Normalized Root Mean Squared Error, $\textrm{NRMSE} = \sqrt{\langle \lVert \mathbf{y}[t] - \hat{\mathbf{y}}[t] \rVert^2 \rangle / \langle \lVert \hat{\mathbf{y}}[t] - \langle\mathbf{y}[t]\rangle \rVert^2 \rangle},$ where $\hat{\mathbf{y}}[t]$ is the prediction provided by the ESN and $\mathbf{y}[t]$ is the desired/teacher output.
The Prediction accuracy is defined as $\gamma=\max\{1-\mathrm{NRMSE}, 0\}$.

In the following, we describe the datasets used in this experimental campaign.

\textbf{Sinusoidal input:} we feed an ESN with a sinusoid $y(t) = \mathrm{sin}( \psi t)$ and we predict future input values with a forecast step $\tau_f = 2 \pi/\psi$.

\textbf{Mackey-Glass time series:} the Mackey-Glass (MG) system is commonly used as a benchmark in chaotic time series prediction. The input signal is generated from the MG time-delay differential equation
\begin{equation}
\label{eq:MG_signal}
\frac{dx}{dt} = \frac{\alpha x(t-\tau_{\mathrm{MG}})}{1+ x(t-\tau_{\mathrm{MG}})^{10}} - \beta x(t).
\end{equation}
We adopt the standard parameters  $\tau_{\mathrm{MG}} = 17, \alpha = 0.2, \beta = 0.1$, initial condition $x(0)=1.2$, and integration step equal to 0.1. The forecast step here is $\tau_f = 6$.

\textbf{Multiple superimposed oscillator:} prediction of superimposed sinusoidal waves with incommensurable frequencies is a hard forecasting exercise, due to the extension of the wavelength \cite{jaeger2004harnessing}.
The ESN is fed with the multiple superimposed oscillator (MSO)
\begin{equation}
\label{eq:mso_signal}
y(t) = \mathrm{sin}(0.2t) + \mathrm{sin}(0.311t) + \mathrm{sin}(0.42t).
\end{equation}
For this task, the ESN is trained to predict future input values, with forecast horizon $\tau_f = 16$.

\textbf{NARMA:} the chosen Non-Linear Auto-Regressive Moving Average (NARMA) task\cite{jaeger2002adaptive} consists in modeling the output of the $r$-order system:
\begin{equation}
y[t + 1] = 0.3y[t] + 0.05y[t]\left(\sum \limits_{i=0}^{r} y[t - i]\right) + 1.5x[t - r]x[t] + 0.1. 
\end{equation}
$x[t]$ is a uniform random noise signal in $[0, 1]$ and is the input of the ESN, which is trained to reproduce $y[t+1]$.
The NARMA task is known to require a memory of at least $r$ past time-steps, since the output is determined by inputs and outputs from the last $r$ time-steps. For this task we set $r=20$ and $\tau_f=15$.

\textbf{Polynomial task:} the ESN is fed with uniform noise in $[-1,1]$ and is trained to reproduce the following output
\begin{equation}
 \label{eq:poly}
 y[t] = \sum \limits_{i=0}^p \sum \limits_{i=0}^{p-i} c_{i,j} x^i[t]x^j[t-d] \;\;\; \text{s.t.} \ \; i + j \leq p,
\end{equation}
where $c_{i,j}$ is uniformly distributed in $[0, 1]$ \cite{Butcher201376}. The difficulty of prediction can be controlled by varying the polynomial degree $p$ and the time delay $d$. For this task we set $p=7$, $d=10$, and $\tau_f=d$.

\begin{figure}[!ht]
  \centering
  \includegraphics[width=0.3\textwidth,keepaspectratio]{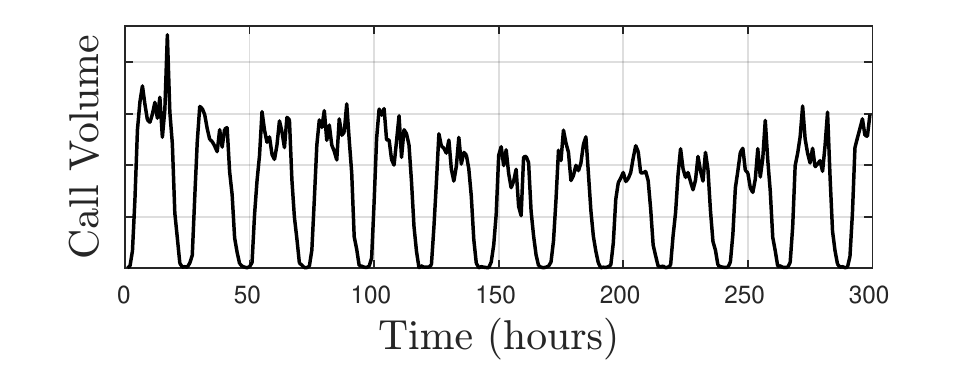}
  \caption{D4D dataset -- load profile of incoming calls volume for the first 300 time intervals.}
  \label{fig:callVolume}
\end{figure}

\textbf{Telephone call load time series:}
as a last test, we consider a real-world dataset relative to the load of phone calls registered over a mobile network.
The data comes from the Orange telephone dataset, published in the Data for Development (D4D) challenge \cite{DBLP:journals/corr/abs-1210-0137}.
D4D is a collection of call data records, containing anonymized events of Orange's mobile phone users in the Ivory Coast, in a period spanning from December 1, 2011 to April 28, 2012.
The dataset consists of 6 time series consisting of: number and volume of incoming calls, number and volume of outgoing calls, day and time (1 hour resolution) when the telephone activity was registered.
More detailed information is available at the related website \cite{D4Dwebsite}. 
All 6 time series are fed into the ESN as inputs; the goal is to predict 6 hours ahead the volume of incoming calls -- the profile of this latter time series is depicted in Fig. \ref{fig:callVolume}. 
%

In each test, we evaluate the correlation between the average entropy of vertex properties in the multiplex and the prediction accuracy $\gamma$, as we vary the hyperparameters $\rho$ and $\omega_i$.
Multiplexes are generated using both the binary and weighted version of the HVG adjacency matrix.
To appreciate the effectiveness of our methodology, we also estimate the correlations of $\gamma$ with $\lambda$, the minimum singular value of the Jacobian of the reservoir (see Methods).
Additionally, we consider the correlation of $\gamma$ with the two layer-based measures IMI and AEO (see Methods).
Correlations are evaluated as follows.
For each configuration $(\rho=k, \omega_i=j)$, we have the prediction accuracy $\gamma_{k,j}$, the entropy $\bar{H}^\text{DG}_{k,j}$, and so on. 
The values assumed by these quantities by varying $\rho$ and $\omega_i$ generate a two-dimensional manifold. 
The point-wise linear (Pearson) correlation between the manifolds relative to $\gamma$ and the other considered measures is the result we are interested in.
The configurations of the hyperparameters that are examined are generated by varying $\rho$ in $[0.5,1.3]$ (20 different values) and $\omega_i$ in $[0.2,0.9]$ (10 different values). 
A total of 200 configurations are evaluated.
Due to the stochastic nature of the ESN initialization, for each configuration $(\rho=k, \omega_i=j)$ we compute $\gamma_{k,j}$ and all the other measures 15 different times.
Successively, we compute the correlations among their average values.
We use a reservoir with $N_r=100$ neurons and sparsity of the internal connectivity equal to $25\%$.
The readout is trained by standard ridge regression with regularization parameter set to $0.05$. The distributions of the vertex properties are estimated by histograms with $b=50$ bins. 
We comment that the reservoir size $N_r$ can be increased/decreased without affecting the applicability of the proposed methodology (only the number of bins used to estimate the vertex properties distribution might be modified -- see Methods).
However, the number of neurons has an impact on the overall performance of the network and on the time and space complexities of the proposed method.

\begin{figure}[!ht]
\centering
  \includegraphics[width=\textwidth,keepaspectratio]{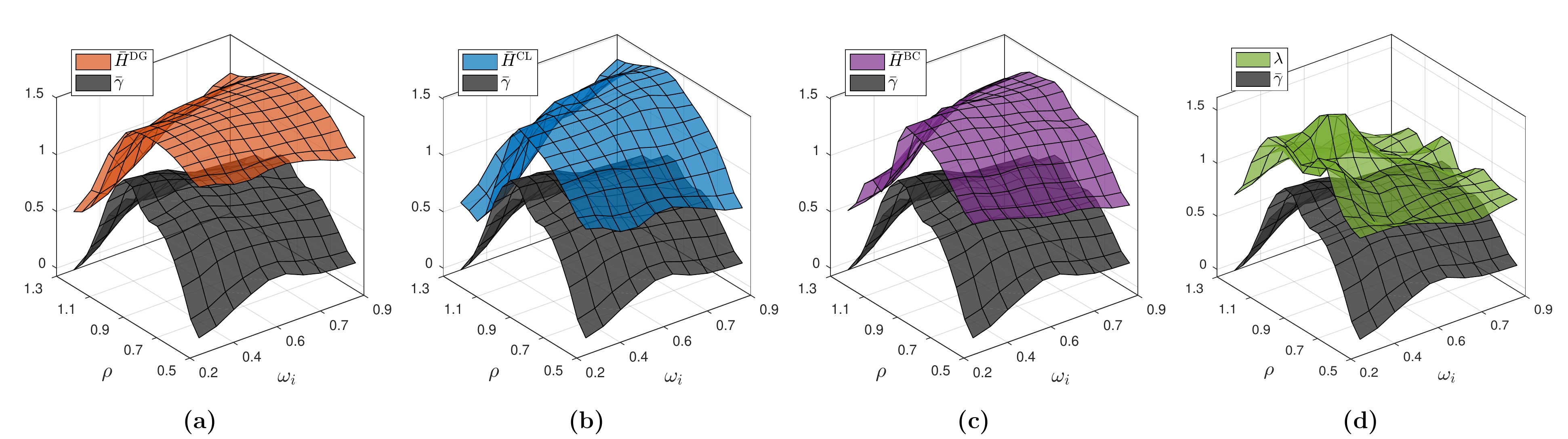}
  \caption{In each panel, the two-dimensional gray manifold represents the values of the prediction accuracy $\gamma$ in the MSO task, for different configurations of $\rho$ and $\omega_i$.
  The colored manifolds are the average entropy of the distributions of (a) the vertex degree $\bar{H}^\text{DG}$, (b) the clustering coefficient $\bar{H}^\text{CL}$, (c) the betweenness centrality $\bar{H}^\text{CC}$. The manifold in (d) represents $\lambda$, which is computed from the ESN Jacobian. From the figure, we observe strong correlations between supervised prediction accuracy and the unsupervised measures (a--c) derived from the multiplex representation of the ESN activity.}
\label{fig:manifolds}
\end{figure}

Fig. \ref{fig:manifolds} depicts the values assumed by $\gamma$ and four graph-based measures in the case of the MSO prediction task.
As can be seen, high correlation emerges between the (average) entropy of the vertex properties and the prediction accuracy.
Since our approach is fully unsupervised, the proposed graph-based measures can approximate well the accuracy $\gamma$, regardless of the task learned by the readout (prediction, function approximation, reconstruction of past inputs, etc).
In Tab. \ref{tab:correlations}, we report the average correlation values and their statistical significance (expressed by $p$-values) on all tasks.
As we can see, the highest (and statistically significant) correlation is achieved by using one of the four average entropy measures of vertex properties.
In particular, the measure based on vertex cluster coefficient distribution, $\bar{H}^\text{CL}$, achieves the best results in 5 of the 6 tasks.
For what concerns the D4D time series, we observe that $\lambda$ achieves high correlation with $\gamma$, but still lower than the one achieved by $\bar{H}^\text{CL}$.
This demonstrates the effectiveness of the proposed methodology, also in the case of a real-world application.
In SIN and MSO tasks, the graph-based quantifiers estimated on the weighted HVG achieve a higher degree of accuracy, with respect to the binary counterpart. 
In these cases, additional qualitative information relative to temporal and amplitude differences in the connected data allows to better represent the dynamics of the system.
Finally, it is worth noting that IMI takes high, yet negative correlation values on both MG and POLY tasks. 
In such cases, results are close to the ones achieved with our approach.

\bgroup
\def\arraystretch{1.1} 
\setlength\tabcolsep{.35em} 
\begin{center}
\begin{table}[!ht]\scriptsize
  \centering
  \caption{Correlations and related $p$-values (in brackets) of the proposed graph-based measures with accuracy $\gamma$ in different prediction tasks, as $\rho$ and $\omega_i$ change. We report the correlation of the manifolds generated by the values of $\gamma$ with the manifolds relative to $\bar{H}^\text{DG}$, $\bar{H}^\text{CL}$, $\bar{H}^\text{BC}$, and $\bar{H}^\text{CC}$, which are the average entropy values of the distributions of vertex degree, clustering coefficient, betweenness and closeness centrality. Each measure is computed on both the binary (b) and weighted (w) versions of the HVG adjacency matrix (adj).
  We also report the correlations of $\gamma$ with the manifolds relative to the minimum singular value of the reservoir Jacobian over time ($\lambda$) and the two multiplex-based measures AEO and IMI, presented in \cite{lacasa2015network}. In each task, the highest correlations with $\gamma$ are highlighted in bold.}
  \vspace{0.2cm}
  \begin{tabular}{r|ccccc|cc|c}
    \cmidrule[1.5pt]{1-9}
    \textbf{Task} & adj & $\mathrm{corr}\left(\gamma,\bar{H}^\text{DG}\right)$ & $\mathrm{corr}\left(\gamma,\bar{H}^\text{CL}\right)$ & $\mathrm{corr}\left(\gamma,\bar{H}^\text{BC}\right)$ & $\mathrm{corr}\left(\gamma,\bar{H}^\text{CC}\right)$ & $\mathrm{corr}\left(\gamma,\mathrm{IMI}\right)$ & $\mathrm{corr}\left(\gamma,\mathrm{AEO}\right)$ & $\mathrm{corr}\left(\gamma,\lambda\right)$ \\
    \cmidrule[1.5pt]{1-9}
    \multirow{2}{*}{SIN} & b & 0.489 (0.006) & 0.488 (0.006) & 0.157 (0.333) & 0.042 (0.797) & \multirow{2}{*}{ -0.091 (0.632)} & \multirow{2}{*}{-0.326 (0.040)} & \multirow{2}{*}{0.154 (0.343)} \\
                         & w & 0.662 (0.000) & \textbf{0.705 (0.000)} & 0.694 (0.000) & -0.127 (0.436) & & & \\
    \cmidrule[.5pt]{1-9}
    \multirow{2}{*}{MG} & b & 0.577 (0.000) & \textbf{0.652 (0.000)} & -0.37 (0.019) & 0.438 (0.005) & \multirow{2}{*}{-0.617 (0.000)} & \multirow{2}{*}{0.414 (0.008)} & \multirow{2}{*}{-0.19 (0.239)} \\
                        & w & -0.138 (0.396) & 0.330 (0.038) & 0.046 (0.777) & 0.564 (0.000) & & & \\
    \cmidrule[.5pt]{1-9}
    \multirow{2}{*}{MSO} & b & -0.215 (0.183) & -0.206 (0.201) & 0.427 (0.006) & 0.333 (0.036) & \multirow{2}{*}{-0.238 (0.139)} & \multirow{2}{*}{-0.312 (0.05)} & \multirow{2}{*}{0.571 (0.000)} \\
                         & w & 0.628 (0.000) & 0.810 (0.000) & \textbf{0.820 (0.000)} & -0.246 (0.125) & & & \\
    \cmidrule[.5pt]{1-9}
    \multirow{2}{*}{NARMA} & b & 0.511 (0.001) & \textbf{0.514 (0.001)} & -0.332 (0.037) & -0.473 (0.002) & \multirow{2}{*}{-0.543 (0.000)} & \multirow{2}{*}{-0.472 (0.002)} & \multirow{2}{*}{0.399 (0.011)} \\
                           & w & -0.373 (0.018) & -0.185 (0.254) & -0.420 (0.007) & -0.376 (0.017) & & & \\
    \cmidrule[.5pt]{1-9}
    \multirow{2}{*}{POLY} & b & 0.755 (0.000) & \textbf{0.765 (0.000)} & -0.393 (0.012) & -0.306 (0.055) & \multirow{2}{*}{-0.745 (0.000)} & \multirow{2}{*}{0.440 (0.005)} & \multirow{2}{*}{-0.557 (0.000)} \\
                          & w & -0.133 (0.412) & -0.113 (0.487) & 0.47 (0.002) & -0.171 (0.291) & & &  \\           
    \cmidrule[.5pt]{1-9}
    \multirow{2}{*}{D4D} & b & 0.632 (0.000) & \textbf{0.677} (0.000) & -0.233 (0.104) & 0.061 (0.676) & \multirow{2}{*}{-0.611 (0.000)} & \multirow{2}{*}{0.168 (0.243)} & \multirow{2}{*}{0.670 (0.000)} \\
                          & w & -0.455 (0.001) & -0.604 (0.000) & 0.189 (0.188) & -0.409 (0.003) & & &  \\   
    \cmidrule[1.5pt]{1-9}
  \end{tabular}
  \label{tab:correlations}
\end{table}
\end{center}
\egroup

\subsection*{Test for memory capacity}

The performed experiment consists in generating 100 different random reservoirs, each one characterized by an increasing value of spectral radius $\rho$ in the $[0.1, 2]$ interval.
As $\rho$ varies, we evaluate the MC by training four readouts in order to reproduce different time-lagged versions of input signal $\mathbf{x}_{10,5}, \dots, \mathbf{x}_{25,20}$. 
Then, on the output of each reservoir, we evaluate the similarities $\delta^\text{TS}$, $\delta^\text{DG}$, and $\delta^\text{AND}$, which are high if there exists at least one series of activations that is similar to the considered past input sequence. 
This is evaluated in such as way that the measure $\kappa^*$ taken into account.
Even if some neurons retain dynamics of previous input sequences, the reservoir introduces shifts in the phase and the amplitude of the input signal.
To filter out these differences, in this test we consider only HVGs defined by binary adjacency matrices, which do not account for differences in the amplitude of the connected values.
To evaluate the effectiveness of the proposed unsupervised memory measures, we compute the correlation between the supervised MC and $\delta^\text{TS}$, $\delta^\text{DG}$, $\delta^\text{AND}$, as $\rho$ varies within the chosen interval.
Note that we only monitor the effect of $\rho$ on the dynamics, since it is the hyperparameter that mostly affects the memory capacity \cite{ozturk2007analysis}. 
We kept the input scaling fixed, $\omega_i = 0.7$, while the remaining hyperparamers are configured as in the previous experiment. 
As before, we repeated each experiment 15 times with different and independent random initializations.
In Tab. \ref{tab:mem_res}, we show the mean correlation values, along with the standard deviations, between the MC and the proposed unsupervised measures of memory capacity.
%
\bgroup
\def\arraystretch{1.1} 
\setlength\tabcolsep{.35em} 
\begin{center}
\begin{table}[!ht]\scriptsize
  \centering
  \caption{Mean correlations and standard deviations of MC with the unsupervised memory quantifiers $\delta^{\text{TS}}$, $\delta^{\text{DG}}$, and $\delta^{\text{AND}}$. We consider 4 different sequences of past inputs $\mathbf{x}_{10,5}$, \dots, $\mathbf{x}_{25,20}$. Values for each measure are computed as the spectral radius $\rho$ of the ESNs reservoirs varies from 0.1 to 2. To compute $\delta^{\text{TS}}$ and $\delta^{\text{DG}}$, we consider three different similarities: the Pearson correlation $\kappa^{\mathrm{PC}}$, the Spearman correlation $\kappa^{\mathrm{SC}}$, and the mutual information $\kappa^{\mathrm{MI}}$. Best results for each input sequence are reported in bold.}
  \vspace{0.2cm}
  \begin{tabular}{c|ccc|ccc|c}
    \cmidrule[1.5pt]{1-8}
    \textbf{Input} & \multicolumn{3}{c|}{$\textbf{corr}\left(\delta^{\text{TS}}, \text{MC}\right)$} & \multicolumn{3}{c|}{$\textbf{corr}\left(\delta^{\text{DG}}, \text{MC}\right)$} & $\textbf{corr}\left(\delta^{\text{AND}}, \text{MC}\right)$ \\
    \textbf{sequence} & $\kappa^{\mathrm{PC}}$ & $\kappa^{\mathrm{SC}}$ & $\kappa^{\mathrm{MI}}$ & $\kappa^{\mathrm{PC}}$ & $\kappa^{\mathrm{SC}}$ & $\kappa^{\mathrm{MI}}$ & \\
    
    \cmidrule[1.5pt]{1-8}
    $\mathbf{x}_{10,5}$ & $0.608 \pm 0.094$ & $0.594 \pm 0.096$ & $0.235 \pm 0.054$ & $0.682 \pm 0.036$ & $0.693 \pm 0.029$ & $0.607 \pm 0.057$ & $\mathbf{0.771 \pm 0.034}$ \\
    
    \cmidrule[.5pt]{1-8}
    $\mathbf{x}_{15,10}$ & $0.542 \pm 0.042$ & $0.546 \pm 0.048$ & $0.484 \pm 0.047$ & $0.547 \pm 0.052$ & $\mathbf{0.645 \pm 0.038}$ & $0.343 \pm 0.056$ & $0.518 \pm 0.043$ \\
    
    \cmidrule[.5pt]{1-8}
    $\mathbf{x}_{20,15}$ & $0.556 \pm 0.068$ & $0.550 \pm 0.065$ & $0.387 \pm 0.038$ & $0.776 \pm 0.040$ & $\mathbf{0.818 \pm 0.046}$ & $0.182 \pm 0.068$ & $0.665 \pm 0.075$ \\
    
    \cmidrule[.5pt]{1-8}
    $\mathbf{x}_{25,20}$ & $0.607 \pm 0.043$ & $0.603 \pm 0.050$ & $0.431 \pm 0.066$ & $0.811 \pm 0.015$ & $\mathbf{0.828 \pm 0.019}$ & $0.501 \pm 0.052$ & $0.468 \pm 0.147$ \\
                   
    \cmidrule[1.5pt]{1-8}
  \end{tabular}
  \label{tab:mem_res}
\end{table}
\end{center}
\egroup

\begin{figure}[!ht]
\centering
  \includegraphics[width=0.55\textwidth,keepaspectratio=true]{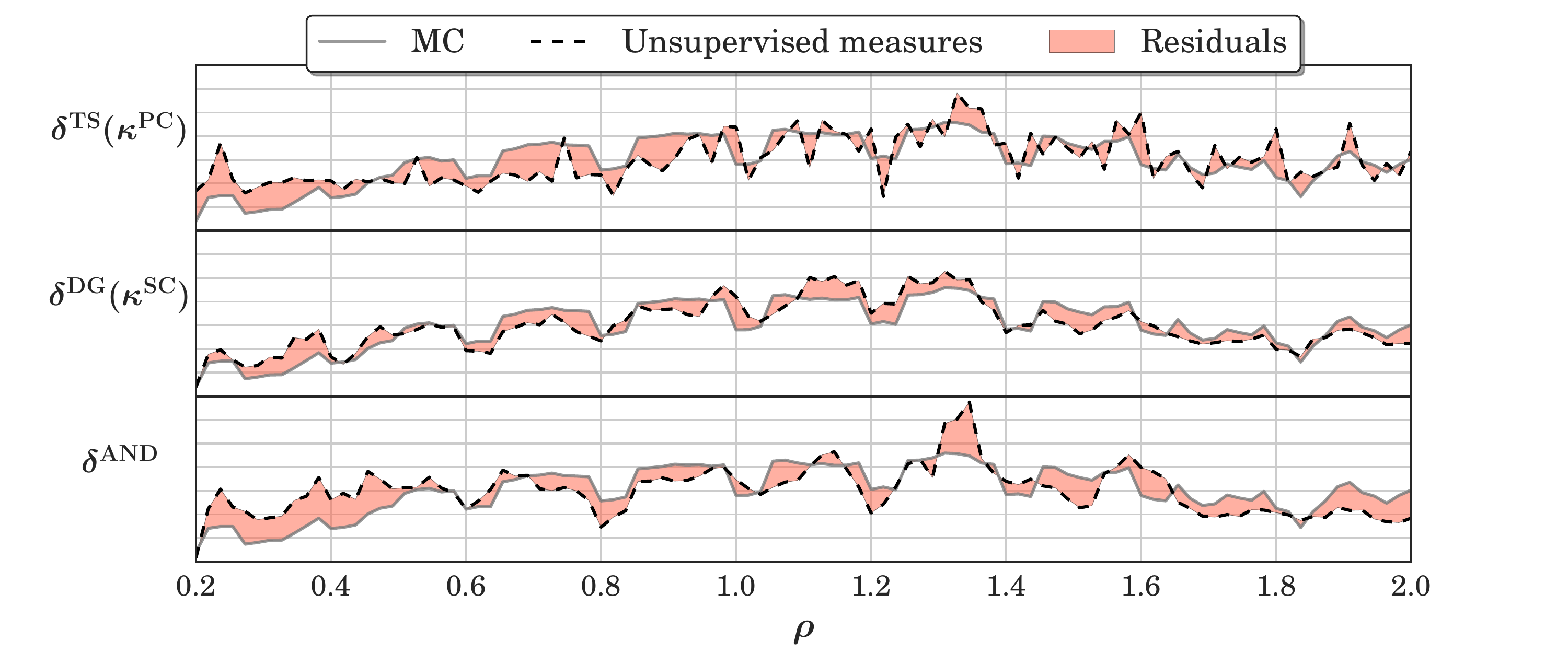}
  \caption{Values of supervised MC and of two selected unsupervised memory measures, when the input sequence $\mathbf{x}_{20,15}$ is taken into account. 
  The smaller the residuals (red areas), the better.
  We can observe that the agreement of the MC with $\delta^\mathrm{DG}$ (using $\kappa^\mathrm{SC}$) is higher than $\delta^\mathrm{TS}$ (using $\kappa^\mathrm{PC}$) and $\delta^\mathrm{AND}$, as $\rho$ varies in $[0.2, 2]$.}
\label{fig:mem_res}
\end{figure}

From the table, we observe that the best agreement with the MC is achieved by the measures derived from the HVGs.
In particular, $\delta^\mathrm{DG}$ configured with the Spearman rank $\kappa^\mathrm{SC}$ is always highly correlated with MC and, in three of the four delayed input sequences taken into account, is the best performing one.
In each setup, $\delta^\mathrm{DG}$ works better if configured with $\kappa^\mathrm{SC}$ rather than $\kappa^\mathrm{PC}$. 
Instead, results obtained with $\kappa^\mathrm{MI}$ are significantly worse in all cases.
$\delta^\mathrm{AND}$ achieves the best results only for the first time lag taken into account, while the agreement with MC is lower in the remaining cases.
Interestingly, several measures show a high degree of correlation with the MC as the size of the delay increases.
$\delta^\mathrm{TS}$, the unsupervised measure computed directly on the input time series and neuron activations, shows positive correlations with the MC, but the agreement is always lower with respect to the graph-based measures.
For $\delta^\mathrm{TS}$, the setting with $\kappa^\mathrm{PC}$ works better than $\kappa^\mathrm{SC}$.
Finally, also in this case by using $\kappa^\mathrm{MI}$ we obtain the worst performance.
In Fig. \ref{fig:mem_res}, we show an example of the values of MC, $\delta^\mathrm{TS}$ (configured with $\kappa^\mathrm{PC}$), $\delta^\mathrm{DG}$ (configured with $\kappa^\mathrm{SC}$), and $\delta^\mathrm{AND}$, as $\rho$ is varied within the $[0.2, 2]$ interval.
%

\section*{Discussion}
\label{sec:discussion}

Experimental results show satisfactory correlations for average entropy of vertex properties with respect to prediction accuracy.
Moreover, the two unsupervised graph-based memory measures that we proposed ($\delta^\mathrm{DG}$ and $\delta^\mathrm{AND}$) correlate well with the supervised measure of memory capacity.

We first discuss the results of the prediction accuracy test, where we analyzed topological properties of vertices in the multiplex, representing the ESN instantaneous state.
On all tests taken into account, we observed a remarkable correlation between $\gamma$ and the average entropy of the clustering coefficient distribution $\bar{H}^\text{CL}$, hence suggesting that the clustering coefficient is able to describe well the heterogeneity of the activations.
To explain this result, it is necessary to elaborate on the properties of the clustering coefficient $\mathrm{CL}(\cdot)$. 
In the HVG literature, $\mathrm{CL}(\cdot)$ its behavior has been analyzed for time series characterized by different Hurst exponent \cite{xie2011horizontal}.
Additionally, an upper bound ($\mathrm{CL}(v) \in [0, 2/\mathrm{DG}(v)]$) is provided for HVGs derived from random time series \cite{PhysRevE.80.046103}. 
In the following, we present an in-depth interpretation of the results by accounting for geometrical properties of the clustering coefficient.

In a HVG, $\mathrm{CL}(v)$ measures the inter-visibility among neighbors of $v$. 
For convex functions, it is possible to connect any two points with a straight line. 
This feature is also (partially) captured by the HVG.
If $v$ is contained in a convex part of the related time series, there is a high degree of intervisibility among the neighbor vertices to which $v$ is connected, hence $\mathrm{CL}(v)$ is high. 
Additionally, moving along the same convex part of the time series results only in minor changes of the clustering coefficient in the associated HVG vertices.
Instead, if $v$ is a local maximum of a concave part, then it is connected to points belonging to two different basins, which do not have reciprocal visibility. 
In this case, $\mathrm{CL}(v)$ is low and its value rapidly changes as one moves away from the maximum. 
This results in great losses of visual information. 
Therefore, large values of $\mathrm{CL}(\cdot)$ indicate the presence of dominating convexities, while low values characterize concavities \cite{costa2007characterization}. 
Accordingly, $\mathrm{CL}(\cdot)$ can be used to measure the length of a convex (concave) part of the time series and how fast the convexity is changing, which is a measure of the fluctuations in the time series \cite{Turner01022001}.
In a regime characterized by contractive dynamics, convexity changes at the same (slow) rate in different neuron activations and this results in a low entropy value of the clustering coefficient distribution among vertices in different layers.
On the edge of criticality, instead, convex and concave parts in the time series of activations are characterized by heterogeneous lengths and they change at different rates. 
This corresponds to a high degree of clustering coefficient diversity of the same HVG vertex, replicated at different layers in the multiplex. 
Finally, in the chaotic regime, all time series fluctuate very rapidly and their convexity changes every few time steps. 
In this case, in each time series of activations the lengths of convex and concave parts are always very short and hence the desired heterogeneity is again lost.
\medskip

For what concerns experiments on memory, the best overall results in terms of agreement with the supervised MC are achieved by the graph-based measure $\delta^\mathrm{DG}$. 
As previously discussed, such a measure evaluates the maximum similarity between the sequence of vertex degrees on the input HVG $G_\mathbf{x}$ and the HVG $G_l$ of neuron activations.
This measure is closely related with the degree distribution $P(k)$, whose importance is known in the HVG literature \cite{PhysRevE.80.046103}.
For example, it has been shown that for time series generated from an i.i.d. process, $P(k)$ follows $P(k) = (1/3)(2/3)^{k-2}$ and the mean degree is $\langle k \rangle=4$. As the correlations in the time series increase, the i.i.d. assumption is lost and $P(k)$ decays faster.
Furthermore, vertex degrees are key parameters to describe dynamic processes on the graph, such as synchronization of coupled oscillators, percolation, epidemic spreading, and linear stability of equilibrium in networked coupled systems \cite{restrepo2007approximating}. Their role has been studied also in the HVG framework \cite{fioriti2012discriminating}.
HVGs have been studied in the context of time series related to processes with power-law correlations \cite{xie2011horizontal}.
In our case, the time series of neuron activations have short-term correlations and increments in the correlation coefficients can have opposite signs at consecutive time lags. 
For these cases, we are not aware of any previous study in terms of HVGs.

In networks which are inherently degree disassortative, the range of degree values increases with network size, with a consequent decrease of the assortativity value \cite{newman2002assortative}. In such networks, the Spearman rank correlation provides a more suitable choice with respect to calculating degree-degree Pearson correlations. It is important to notice that the rank is computed through a non-linear rescaling, which is data dependent. The information on the actual values of the data is discarded as only its inherent ordering (rank) is preserved.
We argue that HVGs convey the same type of information captured by the Spearman correlation. Hence, the latter should be preferred to Pearson correlation to characterize the characteristics of the vertices and related topological properties in HVGs.
This fact justifies the higher agreement with memory capacity achieved by means of $\delta^{\mathrm{DG}}$ when configured with $\kappa^{\mathrm{SC}}$, which accounts for Spearman correlations between sequences of vertex degrees in the HVGs related to the input signal and the neuron activations.
\medskip

Modeling ESN dynamics through a multiplex network allowed us to connect two seemingly different research fields, thus fostering multidisciplinary research in the context of recurrent neural networks. 
By converting a temporal problem into a topological one, we handled temporal dependencies introduced by ESNs (as well as by other types of RNNs), hence overcoming technical limitations of statistical approaches that require independence of samples.
We performed and discussed several experiments that provided empirical evidence that our methodology achieves performance higher than other unsupervised methods and comparable to cross-validation techniques. These results suggest to allocate efforts to further improve the effectiveness of unsupervised learning methods in the context of ESNs and RNNs.
Finally, we would like to stress that, while this paper is primarily focused on network structures in machine learning, our results might suggest new ideas for theoretical understanding of recurrent structures in biological models of neuronal networks \cite{enel2016reservoir,marblestone2016toward,sussillo2015neural}. 
In particular, we believe that it is possible to identify emergent structural patterns in the developed graph-based representations of network dynamics. This would allow to further explore and analyze the route to chaos in input-driven neural models by exploiting the language of graph theory, which is an already established framework within the neuroscience field \cite{sporns2011networks}.

\section*{Additional information}

\textbf{Author contributions} 

\noindent LL outlined the research ideas. FMB and LL conceived methods. FMB performed experiments and generated figures. CA and RJ contributed to the technical discussion. All authors took part to the paper writing and approved the final manuscript.

\noindent \textbf{Competing financial interests} 

\noindent The authors declare that there is no conflict of interest.

\bibliography{Bibliography}

\end{document}